\documentclass{article}

\usepackage[nonatbib,final]{nips}
\usepackage[utf8]{inputenc} 
\usepackage[T1]{fontenc}    
\usepackage{hyperref}       
\usepackage{url}            
\usepackage{booktabs}       
\usepackage{amsfonts}       
\usepackage{nicefrac}       
\usepackage{microtype}      
\usepackage{graphicx}
\usepackage{multirow}
\usepackage[table,xcdraw]{xcolor}
\usepackage{xspace}

\title{CoSformer: Detecting Co-Salient Object with Transformers}

\author{
  Lv Tang \\Nanjing University \\ \texttt{luckybird1994@gmail.com}
  \And Bo Li \\ Independent Researcher \\\texttt{njumagiclibo@gmail.com}
}

\begin{document}
\maketitle

\begin{abstract}

Co-Salient Object Detection (CoSOD) aims at simulating the human visual system to discover the common and salient objects from a group of relevant images. Recent methods typically develop sophisticated deep learning based models have greatly improved the performance of CoSOD task. But there are still two major drawbacks that need to be further addressed, 1) sub-optimal inter-image relationship modeling; 2) lacking consideration of inter-image separability. 
In this paper, 
we propose the Co-Salient Object Detection Transformer (CoSformer) network to capture both
salient and common visual patterns from multiple images. 
By leveraging Transformer architecture, the proposed method address the influence of the input orders and greatly improve the stability of the CoSOD task. 
We also introduce a novel concept of inter-image separability. We construct a contrast learning scheme to modeling the inter-image separability and learn more discriminative embedding space to distinguish true common objects from noisy objects. Extensive experiments on
three challenging benchmarks, i.e., CoCA, CoSOD3k, and
Cosal2015, demonstrate that our CoSformer outperforms cutting-edge models and achieves the new state-of-the-art. We hope that CoSformer can motivate future research for more visual co-analysis tasks.

\end{abstract}

\section{Introduction}
Aiming at simulates the human visual system to discover the common and salient objects from a group of relevant images, co-salient object detection (CoSOD) often serves as a preliminary step for various down-streaming computer vision tasks, e.g., image co-segmentation~\cite{DBLP:conf/cvpr/HsuLC19}, co-localization~\cite{DBLP:conf/cvpr/TangJLF14,DBLP:conf/eccv/JerripothulaCY16} and person re-identiﬁcation~\cite{DBLP:conf/ijcai/LiuZZJ20}. Unlike the standard salient object detection (SOD) which only focuses on the attractive regions from a single image, CoSOD also needs to leverage the similar attributes shared by objects in image group to distinguish the real common objects under the presence of noise objects. 
Although the co-salient objects share the same semantic category, their explicit category attributes are unknown in CoSOD task. That is to say, CoSOD methods are not supposed to model the consistency relations of common objects by using the supervision of specific category labels or other information like temporal relations, which is quite different from video sequences tasks~\cite{DBLP:journals/tip/CongLFPHH19,DBLP:journals/tcsv/JerripothulaCY19}. These unique features make CoSOD an emerging and challenging task which has been rapidly growing in recent few years~\cite{DBLP:journals/corr/abs-2007-03380,DBLP:journals/tcsv/CongLFCLH19,DBLP:journals/tist/ZhangFHBL18}.

Conventional approaches explore the inter-image correlation between image-pairs~\cite{DBLP:journals/tip/LiN11} or a group of relevant images~\cite{DBLP:conf/mm/CaoCTF14} by using constraints or heuristic characteristics like  manifold ranking~\cite{DBLP:journals/tip/CaoTZFF14} and clustering~\cite{DBLP:journals/tip/FuCT13}. 
However, the discrimination of hand-crafted descriptors is too limited to face complex scenes, leading to unsatisfactory performance. Recently deep learning based models have greatly improved the performance of CoSOD task.
By leveraging the Convolutional Neural Networks (CNNs)~\cite{DBLP:conf/ijcai/WeiZBLW17,DBLP:conf/cvpr/ZhangLL019,DBLP:conf/aaai/WangZLX19,DBLP:conf/nips/ZhangCHLZ20,DBLP:conf/nips/Jin0CZG20} and Recurrent Neural Networks (RNNs)~\cite{DBLP:conf/ijcai/0061STSS19,DBLP:conf/iccv/LiSLWH19}, they learn both single image representation (intra-image saliency) and group-wise semantic representation (inter-image consistency) in an end-to-end supervised manner to detect co-salient objects in image group. 
Despite their promising results, we find there are still two major drawbacks that prevent the CoSOD from progressing to the next high level: First, current inter-image relationship modeling is sub-optimal. Second, current methods lack consideration of inter-image separability. 

For the first issue, previous works directly concatenate CNNs features~\cite{DBLP:conf/ijcai/WeiZBLW17} or use RNNs~\cite{DBLP:conf/ijcai/0061STSS19,DBLP:conf/iccv/LiSLWH19} to model the inter-image relationships. However, when assigning different orders of the input images there can output different group representations, which makes both training and inferring procedure unstable. Recent studies ~\cite{DBLP:conf/cvpr/FanLJZFC20,DBLP:conf/nips/ZhangCHLZ20,DBLP:conf/nips/Jin0CZG20} try to alleviate this limitation by applying some sophisticated modification on CNNs architectures. Unfortunately, their efforts do not address the inherent deficiencies of CNNs and RNNs in sequential order modeling. 
To better model the inter-image relationships, we propose to employ the Transformers~\cite{DBLP:conf/nips/VaswaniSPUJGKP17}, which is a widely used sequence to sequence model in Natural Language Processing (NLP) [23]. The self-attention mechanism is designed to learn all pairwise similarities between the input sequence, which empowers Transformers great ability to capture long-range dependencies. Besides, a Transformer model itself is invariant with respect to re-orderings of the input~\cite{DBLP:conf/nips/VaswaniSPUJGKP17,DBLP:journals/corr/abs-2102-11090}. These characteristics of Transformers make them naturally suitable for modeling the inter-image relationships across multiple images. Essentially, intra-image saliency and inter-image consistency are both concerned with relationship modeling: intra-image saliency is to learn the pixel-level relationship within a single image and inter-image consistency is to learn the relationship between images. Thus, we construct Transformer based structures for both intra-image saliency and inter-image consistency modeling. 

\begin{figure}[]
\centering
\includegraphics[scale=0.7]{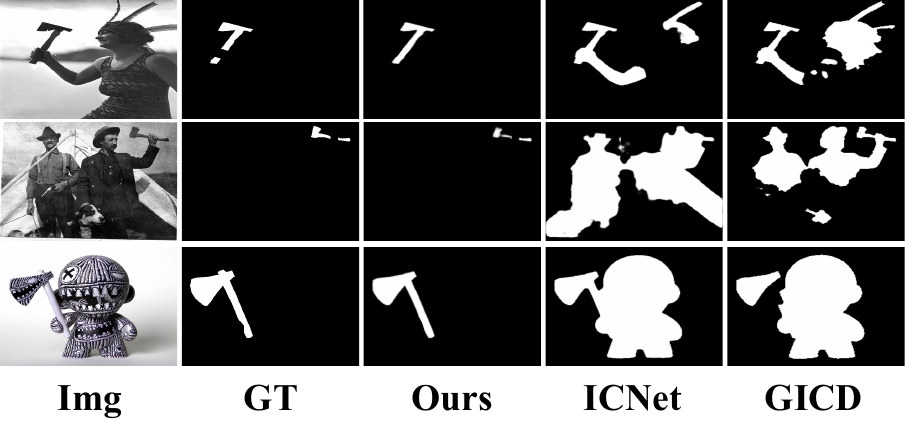}
\caption{Comparison with state-of-the-art methods in complex real-world scenarios.}
\label{introdvis}
\end{figure}

For the second issue, current methods believe that they can well handle the CoSOD task by only using intra-image saliency and inter-image consistency. However, the inter-image consistency only provides positive relations while lacking negative relations between different objects. Training the model only using positive pairs cannot provide enough information for learning a discriminative representation. When facing complex real-world scenarios, the model cannot distinguish true common objects from noisy objects. As can be seen in Fig.\ref{introdvis}, two representative existing methods ICNet~\cite{jin2020icnet} and GICD~\cite{DBLP:conf/eccv/ZhangJXC20} fail to distinguish co-salient objects because they lack consideration of inter-image separability. Inspired by contrastive learning \cite{DBLP:conf/cvpr/He0WXG20}, we propose a novel contrastive loss for CoSOD to model the inter-image separability. We not only regard the co-salient regions in an image group as positive relations but also utilize the non-co-salient regions to build negative relations. Through contrastive loss, the true common objects should be similar to each other and dissimilar to other noisy objects in the embedding space. Finally, we can learn a discriminative representation to get better performance.

In this paper, we propose the Co-Salient Object Detection Transformer (CoSformer) network, which views the CoSOD task as an end-to-end sequence prediction problem. The framework is significantly different from existing approaches. The main contributions can be summarized as follows.

\begin{itemize}
\item CoSformer solves the CoSOD from a new perspective of relationship modeling. By leveraging Transformer architecture, we address the influence of the input orders and greatly improve the stability of deep-based CoSOD methods. Both intra-image saliency and inter-image consistency are naturally modeled by the similar Transformer framework.

\item We provide some insights on the drawbacks of previous methods and proposed a novel concept of inter-image separability. We construct a contrast learning scheme to modeling the inter-image separability and learn more discriminative representations to distinguish true common objects from noisy objects.

\item We validate the performance of our CoSformer on three widely used CoSOD datasets (CoCA, CoSOD3k and Cosal2015), and the performance can outperform other state-of-the-art methods by a large margin. This shows the proposed Transformer framework and contrastive loss can help the network detect a more accurate co-salient result.
\end{itemize}

\section{Related Work}
\textbf{Co-saliency Detection.} The traditional CoSOD methods explore the inter-image correspondence between image-pairs \cite{DBLP:journals/tip/LiN11, DBLP:conf/icip/Chen10a} or a group of relevant images \cite{DBLP:conf/mm/CaoCTF14, DBLP:conf/eccv/JerripothulaCY16} based on shallow handcrafted descriptors \cite{DBLP:conf/cvpr/ChangLL11}. Several studies attempt to capture the inter-image constraints by employing an efficient manifold ranking scheme \cite{DBLP:journals/spl/LiFL015} to obtain guided saliency maps, or using a global association constraint with clustering \cite{DBLP:journals/tip/FuCT13}. 
However, the discrimination of hand-crafted descriptors  are too limited to face the complex scenes, leading to unsatisfactory performance.

Recently deep-based models simultaneously explore the intra-saliency and inter-image consistency in a supervised manner with different approaches, such as graph convolution networks (GCN) \cite{DBLP:conf/mm/0002JZT019, DBLP:conf/cvpr/ZhangLSLC020}, self-learning methods \cite{DBLP:journals/pami/ZhangMH17, DBLP:journals/tnn/ZhangHHS16}, correlation techniques \cite{jin2020icnet}, or co-clustering \cite{DBLP:journals/tip/YaoHZN17}.  Other works explore group-wise semantic representation which is used to detect co-salient regions for each image. There are different methods to capture the discriminative semantic representation, such as group attention semantic aggregation \cite{DBLP:conf/nips/ZhangCHLZ20}, gradient feedback \cite{DBLP:conf/eccv/ZhangJXC20}, recurrent co-attention~\cite{DBLP:conf/ijcai/0061STSS19,DBLP:conf/iccv/LiSLWH19} and even explicit supervision of specific category labels~\cite{DBLP:conf/aaai/WangZLX19,DBLP:conf/nips/Jin0CZG20}. However, most
of the previous methods are unstable during both training and inferring procedure when assigning different orders of the input images. And they all lack consideration of inter-image separability and cannot distinguish true common objects from noisy objects, resulting in ambiguous results when facing complex real-world scenarios. For more about CoSOD tasks, please refer to ~\cite{DBLP:journals/corr/abs-2007-03380,DBLP:journals/tcsv/CongLFCLH19,DBLP:journals/tist/ZhangFHBL18}. Another task related to CoSOD is SOD~\cite{DBLP:journals/corr/abs-1904-09146,Tang_2020_ACCV,DBLP:journals/spm/HanZCLX18,DBLP:conf/cvpr/WeiWWSH020,hu2017online}. For more information about the SOD methods, please refer to survey~\cite{DBLP:journals/tip/BorjiCJL15}.

\textbf{Transformer.} Transformer were first proposed in~\cite{DBLP:conf/nips/VaswaniSPUJGKP17} for the sequence-to-sequence machine translation task, which has revolutionized machine translation and natural language processing.
The Transformer models are then extended to some popular computer-vision tasks including image processing~\cite{DBLP:journals/corr/abs-2012-00364}, object detection~\cite{DBLP:conf/eccv/CarionMSUKZ20}, semantic segmentation~\cite{DBLP:journals/corr/abs-2012-15840}, object tracking~\cite{DBLP:journals/corr/abs-2012-15460}, video instance segmentation~\cite{DBLP:journals/corr/abs-2011-14503}, etc. DETR~\cite{DBLP:conf/eccv/CarionMSUKZ20} builds an object detection system based on Transformers, which largely simplifies the traditional detection pipeline, and achieves on \textit{par} performances compared with highly-optimized CNN based detectors~\cite{DBLP:journals/pami/RenHG017}. ViT~\cite{DBLP:journals/corr/abs-2010-11929} introduces the Transformer to image recognition and models an image as a sequence of patches, which attains excellent results compared to state-of-the-art convolutional networks. The above works show the effectiveness of Transformers in image understanding tasks. More detailed information of the application of the Transformer in the field of computer vision can be found in survey~\cite{DBLP:journals/corr/abs-2012-12556,DBLP:journals/corr/abs-2101-01169}. 

As presented in DETR~\cite{DBLP:conf/eccv/CarionMSUKZ20}, transformer architecture is permutation-invariant, which cannot leverage the order of the tokens in an input sequence. To mitigate this gap, previous works~\cite{DBLP:journals/corr/abs-2010-11929,DBLP:conf/nips/VaswaniSPUJGKP17} add an absolute positional encoding to each token in the input sequence, which enables order-awareness. However, in co-saliency detection task, we want the model should be insensitive to input order when capturing group-wise relationships. Hence, it is natural to use transformer to model group-wise relationships without positional encoding. To our knowledge, thus far there are no prior applications of Transformers to co-saliency detection.

\begin{figure*}[!hbt]
\centering
\includegraphics[scale=0.38]{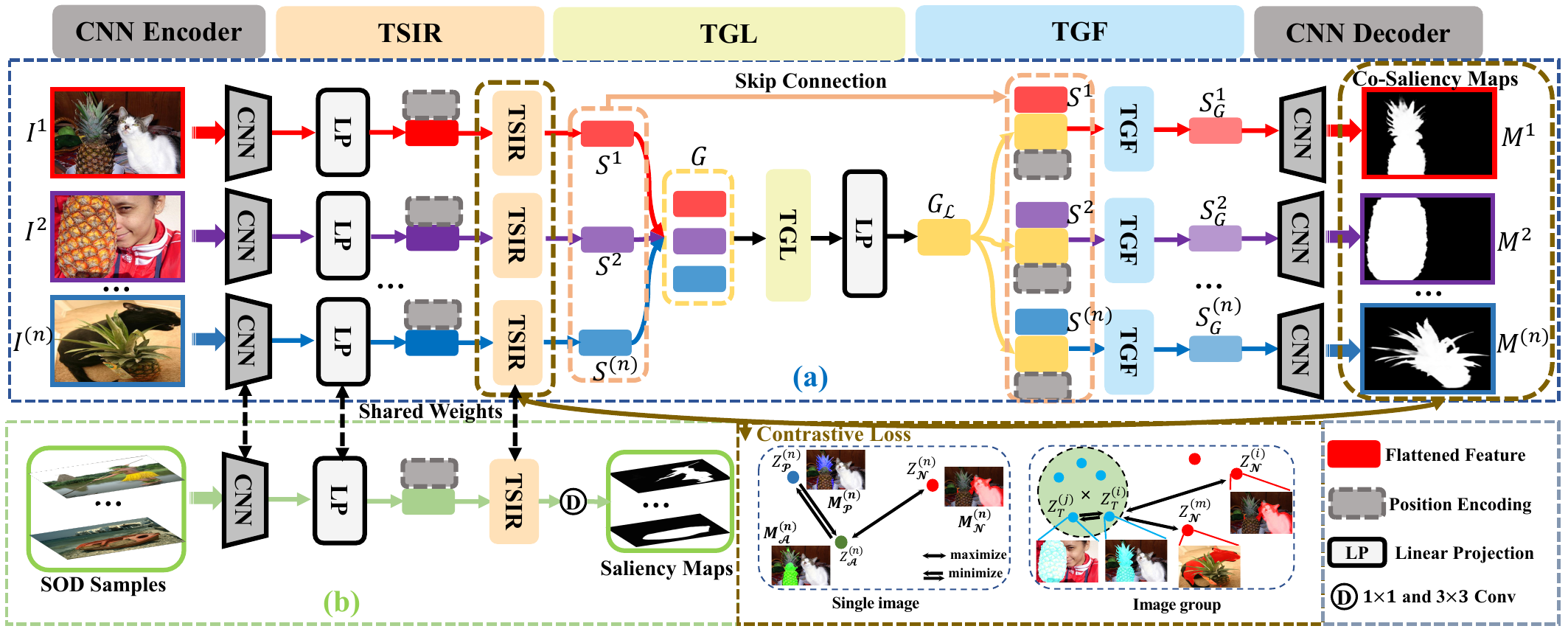}
\caption{The proposed CoSformer framework.}
\label{framework}
\end{figure*}

\section{Proposed Method}
Co-saliency detection aims at discovering the common and salient objects in a group of $N$ relevant images $\mathcal{I}=\{I^{(n)}\}_{n=1}^N$. It is worth raising that directly constructs a pure transformer-based network for co-saliency detection will produce unsatisfactory performance. In SETR~\cite{DBLP:journals/corr/abs-2012-15840}, which is a pure transformer-based network, Transformers treat the input as 1D sequences and exclusively focus on modeling the global context at all Transformer layers, therefore result in low-resolution features which lack detailed low-level information. And this information cannot be effectively recovered by direct upsampling to the full resolution. In co-saliency detection task, Transformers can well model the relation between different pixels, but can not recover fine detailed information. On the other hand, CNN architectures (e.g.,U-Net~\cite{DBLP:conf/miccai/RonnebergerFB15}) provide an avenue for extracting low-level visual cues which can well remedy such fine details. To this end, CoSformer employs a CNN-Transformer architecture to leverage both detailed low-level detailed information from CNN features and the relation encoded by Transformers. We hope that the simplicity of our method will attract new researchers to the co-saliency detection community. Our proposed CoSformer architecture is simple and illustrated in Fig.\ref{framework}.

\subsection{CNN Backbone} 
Starting from the initial image $I^{(n)} \in \mathbb{R}^{H_0 \times W_0 \times 3}$, a conventional CNN backbone (VGG-16~\cite{DBLP:journals/corr/SimonyanZ14a}) generates different levels feature maps $F_l^{(n)} \in \mathbb{R}^{H_l \times W_l \times C_l}$. Following~\cite{DBLP:conf/iccv/ZhaoLFCYC19}, we connect another side path to the last pooling layer in VGG-16, and only use the last four levels features for the following process. For simplicity, these four features can be denoted as a feature set $F^{(n)}$:
\begin{equation}
F^{(n)} = \{F_3^{(n)},F_4^{(n)},F_5^{(n)},F_6^{(n)}\}.
\end{equation}

\subsection{Transformer Encoder}
A good co-saliency detection framework should not only express the intra-saliency of an image, but also reflect the interaction among group images for co-saliency referring. To address these two problems, the proposed Transformer encoder contains three modules: (1) Transformer-based single image representation learning (TSIR) module, which is used to processes each image individually to suppress background noise and capture the saliency of the potential co-salient objects. (2) Transformer-based group representation learning (TGL) module, which can explore all images in the group to learn the inter-image consistency. (3) Transformer-based group fusion (TGF) module, which fuses the learned inter-image consistency and unique intra-image saliency, so the group representation and single saliency representation are sufficiently exploited to facilitate the co-saliency reasoning.

\subsubsection{Preliminary Knowledge}
\begin{figure}[!hbt]
\centering
\includegraphics[scale=0.38]{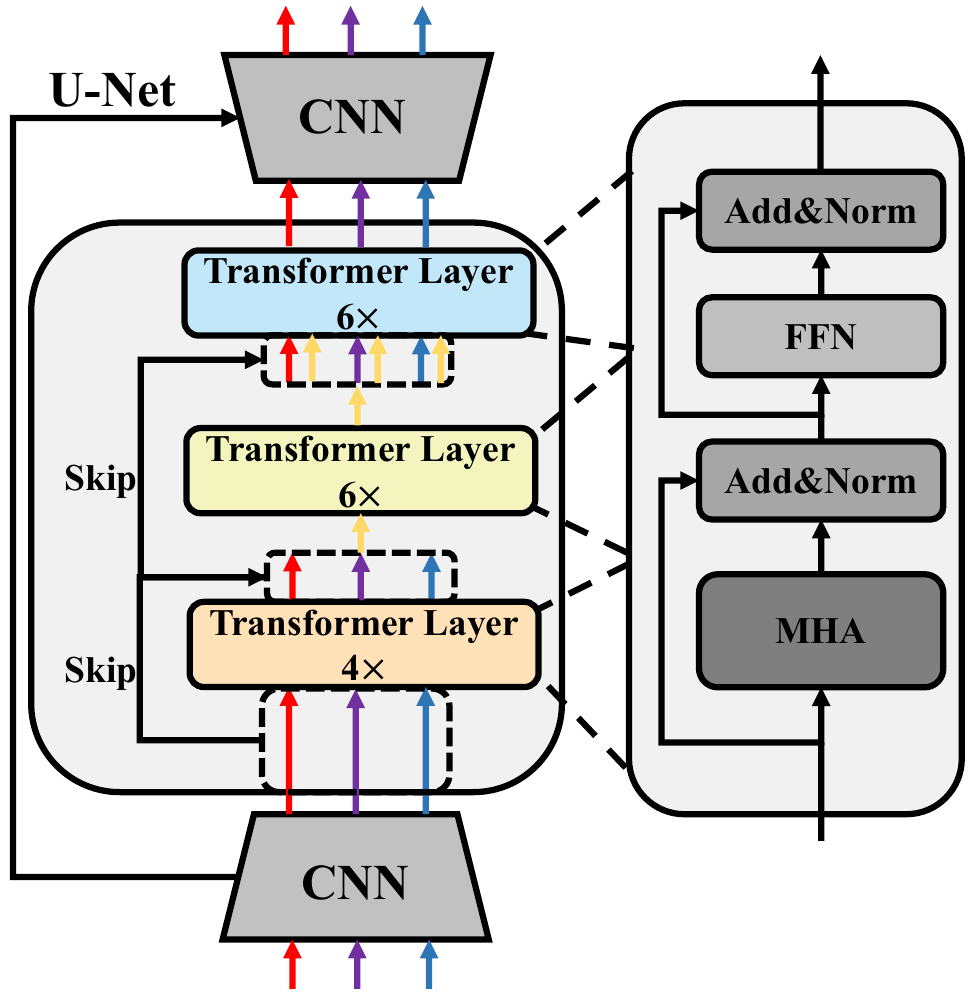}
\caption{The Brief ﬂow chart of the proposed CoSformer.}
\label{U-net}
\end{figure}

Transformer~\cite{DBLP:conf/nips/VaswaniSPUJGKP17} is composed of multi-head attention (MHA) and fully connected feed forward network (FFN). The FFN consists of a $1\times1$ convolution with ReLU activation. Layer Normalization (Norm) is usually added in each MHA and FFN. The structure of a Transformer layer is illustrated in Fig.\ref{U-net}. 

\subsubsection{TSIR Module}
As a basic rule in co-saliency, in most cases, the co-salient regions should be salient with respect to the background in each image. So the network should be able to suppress the background noise and learn the intra-saliency of the potential co-salient objects. To achieve this purpose, the network should consider the relationships between different pixels, then highlights the salient pixels and suppress noisy background pixels. While these pixels will locate in different positions, capturing their long-range dependencies is important. Thus we use Transformer to address this problem.

\textbf{Transformer.} The TSIR module consists of four Transformer layers. First, a $1\times1$ convolution is applied to the $F^{(n)}_{6}$, reducing the dimension from $C_6$ to $d$, resulting in new feature maps $F_6^{(n)} \in \mathbb{R}^{H_6 \times W_6 \times d}$. To form a feature sequence that can be fed into the TSIR, we flatten the spatial dimensions of $F_6^{(n)}$, resulting in a 2D feature map of size $Q\times d$, where $Q=H_6 \times W_6$. While the co-saliency detection task requires position information to locate co-salient objects. To compensate for this, we supplement the features with fixed positional encodings information suggested in~\cite{DBLP:conf/eccv/CarionMSUKZ20}, that contains the two dimensional (horizontal and vertical) positional information. This encoding is added to the input of each MHA.
As can be seen in Fig.\ref{U-net}, we use a skip connection to fuse the output features of TSIR with the previous features. Finally, the fused features can be written as:
\begin{equation}
\mathcal{S}=\{S^{(n)}\}_{n=1}^N,
\end{equation}
where $\mathcal{S} \in \mathbb{R}^{N \times Q \times d}$.

Inspired by CoAD~\cite{DBLP:conf/nips/ZhangCHLZ20}, in the training phase, in addition to the group inputs $\mathcal{I}$ loaded from a CoSOD dataset, we simultaneously feed $K$ auxiliary samples loaded from a SOD dataset into the shared CNN backbone and TSIR, generating single-image saliency maps $\mathcal{H}=\{H^{k)}\}_{k=1}^K$. The saliency and co-saliency prediction is jointly optimized as a multi-task learning
framework with better flexibility and expansibility in terms of providing reliable saliency priors.

\subsubsection{TGL Module}
After TSIR module, the cleaner feature with less background noise will be obtained. Then, the network should capture group-wise relationships to locate co-salient regions. Usually, co-salient objects may be located at different positions across images, so well modeling the relationships between different pixels is difficult for convolution operation. While the self-attention used in Transformer can calculate the relationships between all pixels in group images, which can help model a robust global relationship. it is natural to use Transformer to capture group-wise relationships. Moreover, the learned group-wise relationship is insensitive to the input order of group images. Because the Transformer models the pixel-level relationship among the group features. 

\textbf{Transformer.} To form a group level feature sequence that can be fed into the TGL module, we flatten the first and second dimensions of $\mathcal{S}$ into one dimension, resulting in a group feature sequence $G$ of size $\in \mathbb{R}^{L \times d}$. $L=N \times Q$ is the length of the sequence. The GL module has 6 Transformer layers and each layer consists of an MHA and an FFN. The output of TGL is $G_{\mathcal{L}} \in \mathbb{R}^{L \times d}$, which means the group representation of the image group. We do not add positional encodings on $G$. 

\subsubsection{TGF Module}
As described previously, the group feature is then broadcasted to each individual image, which allows the network to leverage the synergetic information and unique properties between the images. With group representation $G_{\mathcal{L}}$, the network can suppress non-co-salient pixels in $S^{(n)}$ and highlight co-salient pixels.

\textbf{Transformer.} We first use linear projection on $G_{\mathcal{L}}$ to project it to the same size as $S^{(n)}$. The group feature $G_{\mathcal{L}}$ is then broadcasted to each individual image. Taking the concatenation of $S^{(n)}$ and $G_{\mathcal{L}}$ as input, the Transformer decoder outputs feature $S_G^{(n)}$ for each image. The TGF module has 6 Transformer layers and each layer consists of an MHA and an FFN. Like TSIR, We also add fixed positional encodings in each MHA.

\subsection{CNN Decoder}
As the goal of the decoder is to generate the co-saliency results in the original 2D image space ($H_0 \times W_0$), we need to reshape the features from $L \times d$ to a standard 3D feature map $\mathcal{V} \in \mathbb{R}^{N \times H \times W \times d}$. The CNN decoder together with CNN backbone forms a U-shaped architecture that enables feature aggregation at different resolution levels via skip-connections, as shown in Fig.\ref{U-net}. It is worth to be raised that we do not design any extra modules in CNN decoder, so the performance improvement is mainly coming from the proposed Transformer encoder. 

\section{Loss Function}
Inspired by CoAD~\cite{DBLP:conf/nips/ZhangCHLZ20}, we jointly optimize the co-saliency and single image saliency predictions in a multi-task learning framework. Similar to BASNet~\cite{DBLP:conf/cvpr/QinZHGDJ19}, we use pixel-level, region-level and object-level supervision strategy to better keep the uniformity and wholeness of the co-salient objects. Specifically, binary cross-entropy (BCE)~\cite{DBLP:journals/anor/BoerKMR05}, SSIM~\cite{DBLP:journals/tip/WangBSS04} and F-measure  ($F_m$) loss~\cite{DBLP:conf/iccv/ZhaoGWC19} are denoted as pixel-level, region-level and object-level loss. 
we supervise the predicted co-saliency maps $\mathcal{M}=\{M^{(n)}\}_{n=1}^N$ by the corresponding groundtruth $\mathcal{T}=\{{T}^{(n)}\}_{n=1}^N$ under these three loss:
\begin{equation}
L_c = BCE(\mathcal{M},\mathcal{T}) +  SSIM(\mathcal{M},\mathcal{T}) + F_m(\mathcal{M},\mathcal{T}).
\end{equation}

For $K$ auxiliary saliency predictions $\mathcal{H}=\{H^{(k)}\}_{k=1}^K$ , we also supervise them with their groundtruth $\mathcal{T}_s=\{{T_s}^{(k)}\}_{k=1}^K$ under BCE and $F_m$ loss:
\begin{equation}
L_s = BCE(\mathcal{H},\mathcal{T}_s) +  F_m(\mathcal{H},\mathcal{T}_s).
\end{equation}
Because we do not care about the boundary details of $\mathcal{H}$, so we do not need use SSIM loss here. \textit{For the limited space, more details about BCE, SSIM and $F_m$ losses can be found in Supplemental Materials.} 

\subsection{Contrastive Loss}
As described, the purpose of TGF module is to suppress noisy pixels and highlight the remaining co-salient pixels. So we add a novel contrastive loss that can promote the differences between noisy and co-salient pixels, which can help model inter-image separability. The existing contrastive learning methods (e.g. \cite{DBLP:conf/cvpr/He0WXG20, DBLP:conf/icml/ChenK0H20}) are a simple instance discrimination task. It treats each image as an individual instance, and the purpose is that each image can be well distinguished by the contrastive learning framework. Next, we will describe the way to construct positive and negative samples for CoSOD task.

\begin{figure}[!hbt]
\centering
\includegraphics[scale=0.45]{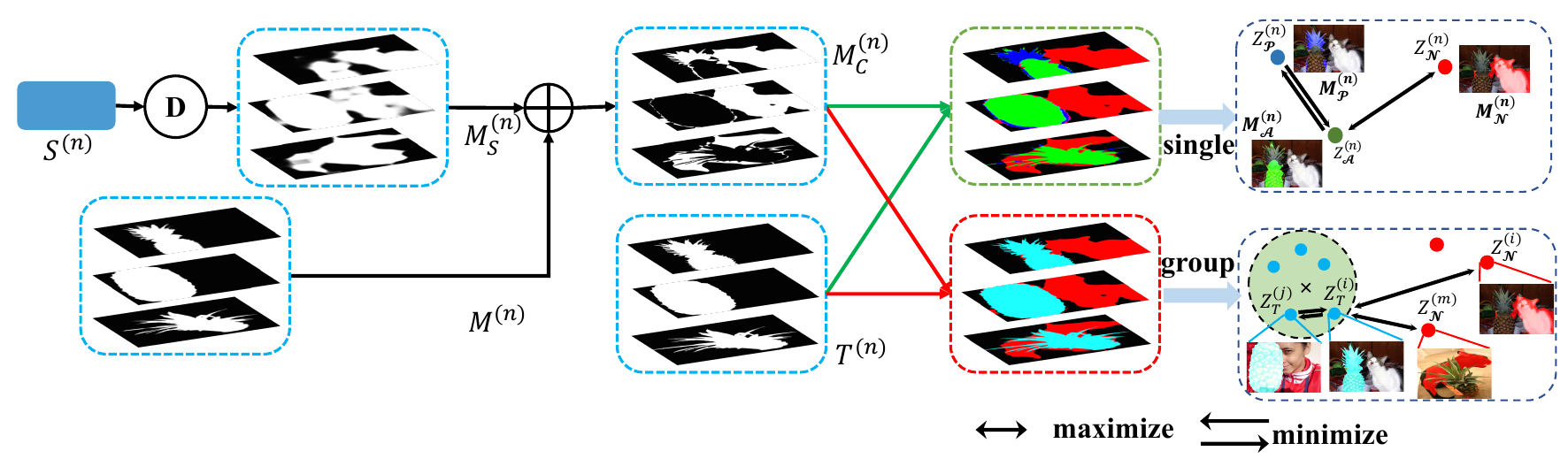}
\caption{Illustration of the proposed contrastive loss.}
\label{Contrastive}
\end{figure}

We first use a $3\times3$ and $1\times1$ convolution on $\{S^{(n)}\}_{n=1}^N$ to obtain co-saliency maps $\{M^{(n)}_S\}_{n=1}^N$, which are supervised by BCE and $F_m$ loss. We denote this loss function as $L_{ct}$. 

Masks $M^{(n)}_S$ and $M^{(n)}$ are binarized with a threshold of 0.5, then we can get a mask by:
\begin{equation}
M^{(n)}_C=M^{(n)}_S\oplus M^{(n)},
\end{equation}
where $\oplus$ means variance operation. Mask $M^{(n)}$ means ours current detected co-saliency map, while $M^{(n)}_S$ means a saliency map that may contain many noisy regions. $M^{(n)}_C$ means the difference between $M^{(n)}_S$ and $M^{(n)}$. \\
\textbf{Contrastive in single image.}  We first compare $M^{(n)}_C$ with groundtruth $T^{(n)}$, and get three masks:
\begin{equation}
M^{(n)}_{\mathcal{A}} = M^{(n)}_{C} \cap T^{(n)}, \\
M^{(n)}_{\mathcal{P}} = T^{(n)} - M^{(n)}_{C}, \\
M^{(n)}_{\mathcal{N}} = M^{(n)}_{C} -T^{(n)},
\end{equation}
where $\cap$ means the intersection operation, and $-$ means the subtraction operation. As can be seen in Fig.\ref{Contrastive}, $M^{(n)}_{\mathcal{N}}$ only contains noisy regions, and $M^{(n)}_{\mathcal{P}}$ and $M^{(n)}_{\mathcal{A}}$ form co-salient regions.

Then we can get corresponding semantic features of $\{ M^{(n)}_{\mathcal{A}},M^{(n)}_{\mathcal{P}},M^{(n)}_{\mathcal{N}}\}$ by multiplying them with feature $S_G^{(n)}$, denoted as $\{ Z^{(n)}_{\mathcal{A}},Z^{(n)}_{\mathcal{P}},Z^{(n)}_{\mathcal{N}}\}$. 
We ensure the integrity of the co-salient objects by minimizing the distance between $Z^{(n)}_{\mathcal{A}}$ and $Z^{(n)}_{\mathcal{P}}$, such as the pineapple and its leaves in Fig.\ref{Contrastive}. On the other hand, we suppress noisy objects by maximizing the distance between $Z^{(n)}_{\mathcal{A}}$ and $Z^{(n)}_{\mathcal{N}}$ like the pineapple and cat in Fig.\ref{Contrastive}. Thus we can construct positive sample pairs $\{Z^{(n)}_{\mathcal{A}}, Z^{(n)}_{\mathcal{P}}\}$ and negative sample pairs $\{Z^{(n)}_{\mathcal{A}}, Z^{(n)}_{\mathcal{N}}\}$ in single image. So the contrastive loss in every single image can be described as:
\begin{equation}
    L_{single} = -\sum^N_{i=1} log\frac{exp(g(Z^{(i)}_{\mathcal{A}}) \cdot g(Z^{(i)}_{\mathcal{P}}) / \tau)}{\sum exp(g(Z^{(i)}_{\mathcal{A}}) \cdot g(Z^{(i)}_{\mathcal{N}}) / \tau)}.
\end{equation}
\textbf{Contrastive in image group.} For CoSOD, the co-salient regions in a image group can be considered as objects which share same semantic. And the indistinguishable regions in different images can help to establish negative relations with co-salient regions. Having more negative samples is crucial for learning good representations. By maximizing the distance between them and the co-salient regions, more constraints can be provided for the co-salient regions, so as to obtain a good embedding space. So we can construct positive sample pairs $\{Z^{(i)}_T,Z^{(j)}_T\}$ and negative sample pairs $\{Z^{(i)}_T,Z^{(m)}_{\mathcal{N}}\}$ in image group, where $i\neq j$ , $i,j,m \in [1, N]$. $Z^{(i)}_T$ can be obtained by multiplying $S^{(i)}_G$ with $T^{(i)}$. Finally, we get the following optimization criterion for image groups:
\begin{equation}
    \centering
    L_{group} = -\sum_i\sum_j log\frac{exp(g(Z^{(i)}_T)\cdot g(Z^{(j)}_T)/\tau)}{\sum_m exp(g(Z^{(i)}_T)\cdot g(Z^{(m)}_{\mathcal{N}})/\tau)}.
\end{equation}

The above $g(\cdot)$ is another non-linear projection head followed \cite{DBLP:conf/icml/ChenK0H20} and the temperature $\tau$ relaxes the dot product. The total contrastive loss can be written as:
\begin{equation}
    L_{cont} = L_{single} + L_{group}.
\end{equation}
Note that all parts of CoSformer are trained jointly, so the over all loss function is given as:
\begin{equation}
L = L_{s} + L_{c} + L_{ct} + L_{cont}.
\end{equation}

\begin{table}[]
\centering
\caption{Quantitative comparison with SOTA on three CoSOD datasets. The best two results are in {\color[HTML]{FF0000} red} , {\color[HTML]{32CB00} green}. Larger $E^{max}_{\phi}$, $S_{\alpha}$, $F_\beta^{max}$, smaller MAE mean better results. }
\scalebox{0.75}{
\begin{tabular}{c|c|cccc|cccc|cccc}
\hline
                          &                        & \multicolumn{4}{c|}{CoCA}                                                                                                 & \multicolumn{4}{c|}{CoSOD3k}                                                                                              & \multicolumn{4}{c}{Cosal2015}                                                                                             \\ \cline{3-14} 
\multirow{-2}{*}{Methods} & \multirow{-2}{*}{Type} & $E^{max}_{\phi}$             & $S_{\alpha}$                 & $F_\beta^{max}$              & MAE                          & $E^{max}_{\phi}$             & $S_{\alpha}$                 & $F_\beta^{max}$              & MAE                          & $E^{max}_{\phi}$             & $S_{\alpha}$                 & $F_\beta^{max}$              & MAE                          \\ \hline
CBCS(TIP2013)             & Co                     & 0.641                        & 0.523                        & 0.313                        & 0.180                        & 0.637                        & 0.528                        & 0.466                        & 0.228                        & 0.656                        & 0.544                        & 0.532                        & 0.233                        \\
GWD(IJCAI2017)            & Co                     & 0.701                        & 0.602                        & 0.408                        & 0.166                        & 0.777                        & 0.716                        & 0.649                        & 0.147                        & 0.802                        & 0.744                        & 0.706                        & 0.148                        \\
RCAN(IJCAI2019)           & Co                     & 0.702                        & 0.616                        & 0.422                        & 0.160                        & 0.808                        & 0.744                        & 0.688                        & 0.130                        & 0.842                        & 0.779                        & 0.764                        & 0.126                        \\
CSMG(CVPR2019)            & Co                     & {\color[HTML]{00B050} 0.735} & 0.632                        & 0.508                        & 0.124                        & 0.804                        & 0.711                        & 0.709                        & 0.157                        & 0.842                        & 0.774                        & 0.784                        & 0.130                        \\
CoEG(TPAMI2020)           & Co                     & 0.717                        & 0.616                        & 0.499                        & {\color[HTML]{00B050} 0.104} & 0.825                        & 0.762                        & 0.736                        & 0.092                        & 0.882                        & 0.836                        & 0.832                        & 0.077                        \\
GICD(ECCV2020)            & Co                     & 0.712                        & {\color[HTML]{00B050} 0.658} & {\color[HTML]{00B050} 0.510} & 0.125                        & 0.831                        & 0.778                        & 0.744                        & 0.089                        & 0.885                        & 0.842                        & 0.840                        & 0.071                        \\
ICNet(NIPS2020)           & Co                     & 0.698                        & 0.651                        & 0.506                        & 0.148                        & 0.832                        & 0.780                        & 0.743                        & 0.097                        & 0.900                        & 0.856                        & 0.855                        & 0.058                        \\
CoAD(NIPS2020)            & Co                     & -                            & -                            & -                            & -                            & {\color[HTML]{00B050} 0.874} & {\color[HTML]{00B050} 0.822} & {\color[HTML]{00B050} 0.786} & {\color[HTML]{00B050} 0.078} & {\color[HTML]{00B050} 0.915} & {\color[HTML]{00B050} 0.861} & {\color[HTML]{00B050} 0.857} & {\color[HTML]{00B050} 0.063} \\
Ours                      & Co                     & {\color[HTML]{FE0000} 0.770} & {\color[HTML]{FE0000} 0.724} & {\color[HTML]{FE0000} 0.603} & {\color[HTML]{FE0000} 0.103} & {\color[HTML]{FE0000} 0.879}    & {\color[HTML]{FE0000} 0.835}    & {\color[HTML]{FE0000} 0.807}    & {\color[HTML]{FE0000} 0.066}    & {\color[HTML]{FE0000} 0.929} & {\color[HTML]{FE0000} 0.894} & {\color[HTML]{FE0000} 0.891} & {\color[HTML]{FE0000} 0.047} \\ \hline
EGNet(ICCV2019)           & Sin                    & 0.631                        & 0.595                        & 0.388                        & 0.179                        & 0.793                        & 0.762                        & 0.702                        & 0.119                        & 0.843                        & 0.818                        & 0.786                        & 0.099                        \\
F3Net(AAAI2020)           & Sin                    & 0.678                        & 0.614                        & 0.437                        & 0.178                        & 0.802                        & 0.772                        & 0.717                        & 0.114                        & 0.866                        & 0.841                        & 0.815                        & 0.084                        \\
MINet(CVPR2020)           & Sin                    & 0.634                        & 0.550                        & 0.387                        & 0.221                        & 0.782                        & 0.754                        & 0.707                        & 0.122                        & 0.847                        & 0.831                        & 0.805                        & 0.181                        \\ \hline
\end{tabular}}
\label{QuantitativeResults}
\end{table}

\begin{figure}[!hbt]
\centering
\includegraphics[scale=0.65]{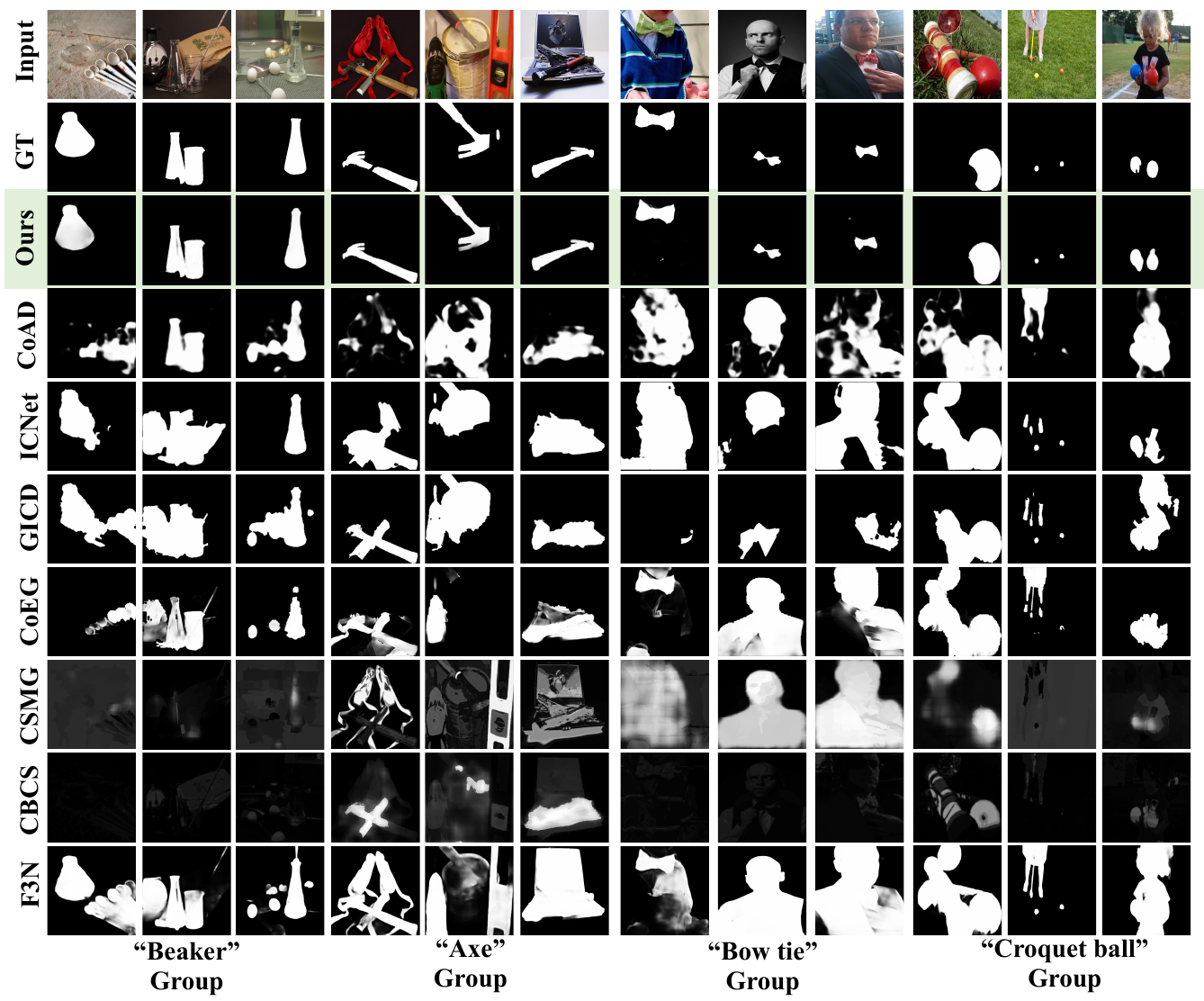}
\caption{Visual comparison between our method and other SOTA methods. It can be clearly observed that our method achieves impressive performance in all these cases.}
\label{QualitativeResults}
\end{figure}

\section{Experiments}
\subsection{Implementation Details}
Following~\cite{DBLP:conf/nips/ZhangCHLZ20,DBLP:conf/eccv/ZhangJXC20,jin2020icnet,deng2020re}, We use VGG-16 as our backbone. The training set is a subset of the COCO dataset~\cite{DBLP:conf/eccv/LinMBHPRDZ14} (9213 images) and saliency dataset DUTS~\cite{DBLP:conf/cvpr/WangLWF0YR17}, as suggested by~\cite{DBLP:conf/nips/ZhangCHLZ20}. In training iteration, 8 ($N=8$) images from a sub-group of COCO dataset and 8 ($K=8$) images from DUTS are simultaneously fed into the network for jointly optimizing. The images are all resized to $256 \times 256$ for training and testing, and the output co-saliency maps are resized to the original size for evaluation. The network is trained over 100 epochs in total with the Adam optimizer. The initial learning rate is set to $1e-4$, $\beta_1=0.9$ and $\beta_2=0.99$.

\subsection{Evaluation Datasets and Metrics}
We employ three challenging datasets for evaluation: CoCA~\cite{DBLP:conf/eccv/ZhangJXC20}, CoSOD3k~\cite{deng2020re}, and Cosal2015~\cite{DBLP:journals/ijcv/ZhangHLWL16}. The last is a large dataset widely used in the evaluation of CoSOD methods. The ﬁrst two were recently proposed for challenging real-world co-saliency evaluation, with the images usually containing multiple common and non-common objects against a complex background. We use maximum E-measure $E^{max}_{\phi}$~\cite{DBLP:conf/ijcai/FanGCRCB18}, S-measure $S_{\alpha}$~\cite{DBLP:conf/iccv/FanCLLB17}, maximum F-measure $F_\beta^{max}$~\cite{DBLP:journals/tip/BorjiCJL15}, and mean absolute error (MAE) to evaluate methods in our experiments. Evaluation toolbox:\url{ https://dpfan.net/CoSOD3K/}.

\subsection{Comparisons with the State-of-the-Arts}
Since not all CoSOD models have publicly released codes or results, we only compare our CoSformer with one representative traditional algorithm (CBCS~\cite{DBLP:journals/tip/FuCT13}) and seven deepbased CoSOD models, including GWD~\cite{DBLP:journals/tip/WeiZBLWZ19}, RCAN~\cite{DBLP:conf/ijcai/0061STSS19}, CSMG~\cite{DBLP:conf/cvpr/ZhangLL019}, CoEG~\cite{deng2020re}, GICD~\cite{DBLP:conf/eccv/ZhangJXC20}, ICNet~\cite{jin2020icnet}, CoAD~\cite{DBLP:conf/nips/ZhangCHLZ20}. We also compare our method with 3 famous single-SOD methods EGNet~\cite{DBLP:conf/iccv/ZhaoLFCYC19}, F3Net~\cite{DBLP:journals/corr/abs-1911-11445} and MINet~\cite{DBLP:conf/cvpr/PangZZL20}. 

\textbf{Quantitative Results.} From Table.\ref{QuantitativeResults}, We can see that compared to other state-of-the-art methods, our model outperforms all of them in all metrics. For example, for dataset CoCA, our method improves the performance by a large margin. Compared to the second ranked performance, the percentage gain reaches 4.7\% for $E^{max}_{\phi}$,  10\% for $S_{\alpha}$, and 18.2\% for $F_\beta^{max}$. On the challenging CoSOD3k and Cosal2015 datasets, our model capitalizes on our better consensus and significantly outperforms other methods. These results demonstrate the efficiency of the proposed CoSformer framework and contrastive loss. The second best method CoAD is established upon the VGG-16 backbone network, containing 121 MB parameters totally. The proposed CoSformer shares a very close number of parameters (115 MB). \textit{For the limited space, P-R curves can be found in Supplemental Materials.} 

\textbf{Qualitative Results.} Fig.\ref{QualitativeResults} shows the co-saliency maps generated by different methods for qualitative comparison. As can be seen, the SOD method F3N can only detect salient objects and fail to distinguish co-salient objects. The CoSOD methods perform better than the SOD methods because of considering group-wise relationships in designing the model. As can be seen in "Beaker Group", these CoSOD can suppress some non-co-salient regions. However, these CoSOD methods only model feature-level group relationships and lack consideration of inter-image separability. When facing complex real-world scenarios, they are unable to handle these challenging cases, like "Axe Group" and "Bow tie Group", where non-co-salient objects are very close to co-salient objects. While our proposed CoSformer models the pixel-level group relationships, and use contrastive loss to model inter-image separability, therefore performs much better on detecting co-salient objects.

\begin{table}[]
\centering
\caption{Ablation studies on the CoSOD3k and Cosal2015 datasets.}
\scalebox{0.85}{
\begin{tabular}{ccccc|cccc|cccc}
\hline
\multicolumn{5}{c|}{Configurations}              & \multicolumn{4}{c|}{CoSOD3k}                                                                                              & \multicolumn{4}{c}{Cosal2015}                                                                                             \\ \hline
Baseline & TSIR    & TGL     & TGF     & Cont    & $E^{max}_{\phi}$             & $S_{\alpha}$                 & $F_\beta^{max}$              & MAE                          & $E^{max}_{\phi}$             & $S_{\alpha}$                 & $F_\beta^{max}$              & MAE                          \\ \hline
$\surd$  &         &         &         &         & 0.785                        & 0.720                        & 0.655                        & 0.144                        & 0.807                        & 0.748                        & 0.710                        & 0.145                        \\
$\surd$  & $\surd$ &         &         &         & 0.803                        & 0.735                        & 0.695                        & 0.128                        & 0.840                        & 0.783                        & 0.767                        & 0.122                        \\
$\surd$  & $\surd$ & $\surd$ &         &         & 0.839                        & 0.784                        & 0.738                        & 0.089                        & 0.880                        & 0.851                        & 0.840                        & 0.078                        \\
$\surd$  & $\surd$ & $\surd$ & $\surd$ &         & 0.860                        & 0.822                        & 0.785                        & 0.071                        & 0.910                        & 0.871                        & 0.875                        & 0.061                        \\
$\surd$  & $\surd$ & $\surd$ & $\surd$ & $\surd$ & {\color[HTML]{FE0000} 0.879} & {\color[HTML]{FE0000} 0.835} & {\color[HTML]{FE0000} 0.807} & {\color[HTML]{FE0000} 0.066} & {\color[HTML]{FE0000} 0.929} & {\color[HTML]{FE0000} 0.894} & {\color[HTML]{FE0000} 0.891} & {\color[HTML]{FE0000} 0.047} \\ \hline
\end{tabular}}
\label{ablation_study}
\end{table}

\begin{figure}[!hbt]
\centering
\includegraphics[scale=0.6]{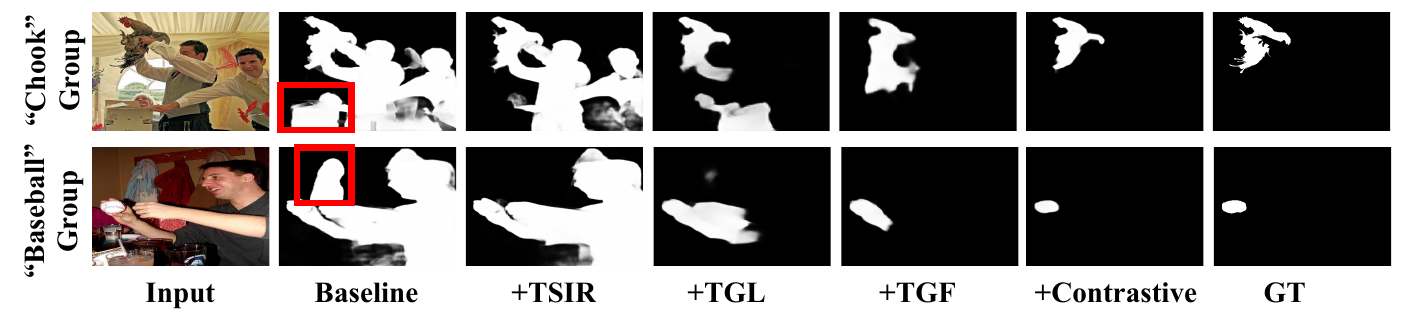}
\caption{Visualization of different ablative results. From left to right: Input image, Co-saliency maps produced by the Baseline, Baseline+TSIR, Baseline+TSIR+TGL, Baseline+TSIR+TGL+TGF and Baseline+TSIR+TGL+TGF+Contrastive Loss}
\label{visablation}
\end{figure}

\begin{figure}[!hbt]
\centering
\includegraphics[scale=0.6]{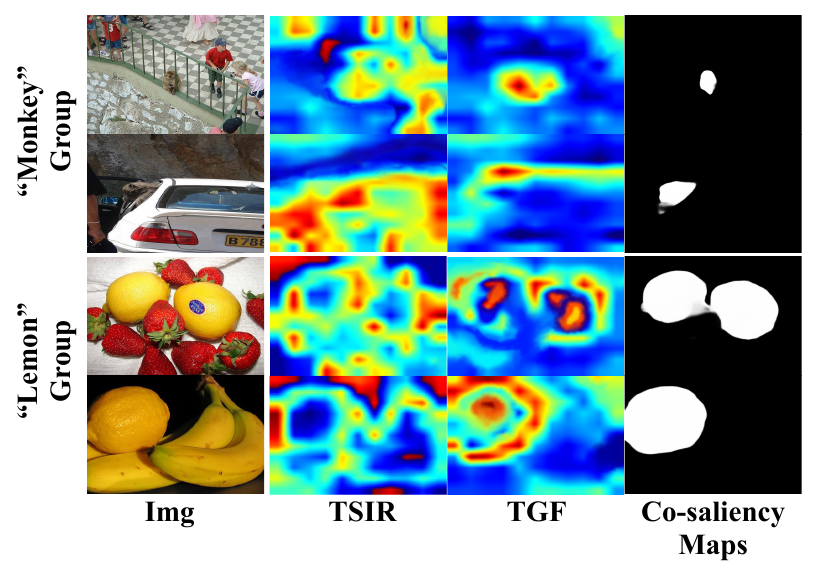}
\caption{Visualization of TSIR and TGF.}
\label{heatmap}
\end{figure}

\subsection{Ablation Studies}
To verify our contributions, we design different variants of our CoSformer with the VGG-16 backbone by replacing the three key modules (\textit{i.e.}, TSIR, TGL and TGF). To construct our baseline model, we simplified the CoSformer as follows: 1) replacing the TSIR module with standard $3\times3$ convolutions; 2) replacing the TGL module with direct concatenation followed by a $1\times1$ convolution; 3) replacing TGF with standard $3\times3$ convolutions layers. We train the baseline model in BCE, SSIM, and $F_m$ losses. The baseline model is carefully designed to share a very similar parameter number with the full model. 

In Fig.\ref{visablation}, it is observed that the baseline model can roughly locate the salient object, but fails to suppress the non-common salient object and background (red boxes). As can be seen in Table.\ref{ablation_study}, By introducing the TSIR, background regions are effectively suppressed, thereby promoting the percentage gain of $F_\beta^{max}$ reaches 6.1\% on CoSOD3k and 4.6\% on Cosal2015. Then, introducing the TGL module that learns more discriminative group semantic representations further suppresses the non-common salient objects, and boosts the performance with large margins. Subsequently, the TGF module can further suppress non-co-salient pixels. As can be seen in Fig.\ref{visablation}, even though TGF can suppress the noise as much as possible, some hard pixels which are close to real co-salient regions are still preserved. Hence, the proposed contrastive loss (Cont) is designed to solve this problem, which can help model inter-image separability. With the contrastive loss, the co-salient objects are highlighted, which further boosts the whole framework to the state-of-the-art on all datasets. 

To get a deeper understanding of the proposed methods, we visualize the learned features from TSIR and TGF in Fig.\ref{heatmap}. The features from TSIR may contain much noise, such as "car", "strawberries". After our proposed pixel-level group relationships modeling and contrastive loss, the features in TGF can be cleaner and focus on co-salient regions.

\section{Conclusion}
Recent methods typically develop sophisticated deep learning based models have greatly improved the performance of CoSOD task. But there are still two major drawbacks that need to be further addressed, 1) sub-optimal inter-image relationship modeling; 2) lacking consideration of inter-image separability. 
In this paper, we propose the Co-Salient Object Detection Transformer (CoSformer) network to capture both
salient and common visual patterns from multiple images. 
By leveraging Transformer architecture, the proposed method address the influence of the input orders and greatly improve the stability of the CoSOD task. 
We also introduce a novel concept of inter-image separability. We construct a contrast learning scheme to modeling the inter-image separability and learn more discriminative embedding space to distinguish true common objects from noisy objects. Extensive experiments on
three challenging benchmarks, i.e., CoCA, CoSOD3k, and
Cosal2015, demonstrate that our CoSformer outperforms cutting-edge models and achieves the new state-of-the-art.

\clearpage

\title{Supplementary Materials for CoSformer: Detecting Co-Salient Object with Transformers}

\author{
  Lv Tang \\Nanjing University \\ \texttt{luckybird1994@gmail.com}
  \And Bo Li \\ Independent Researcher \\\texttt{njumagiclibo@gmail.com}
}

\maketitle

\section{Introduction}
This supplemental material contains three parts:

\begin{itemize}

\item Section 8 gives more quantitative and qualitative experimental results to demonstrate the superiority of our CoSformer.

\item Section 9 gives more details about the BCE, SSIM and $F_m$ losses, and analyzes the role they play in co-saliency detection task. 

\item Section 10 gives more analyses of the proposed TGL module, which further verifies the TGL can make the CoSformer insensitive to the input order of group images.

\end{itemize}
We hope this supplemental material can help you get a better understanding of our work.

\section{More Quantitative and Qualitative Results}

\subsection{Quantitative Comparison on more datasets}
We compare our method with other methods on another two conventional CoSOD datasets iCoseg~\cite{DBLP:conf/cvpr/BatraKPLC10} and MSRC~\cite{DBLP:conf/iccv/WinnCM05}. The results are shown in Table.\ref{table1}. We can see that compared to other state-of-the-art methods, our model outperforms all of them in all metrics. 

\subsection{Qualitative Comparison}
\begin{figure}[!t]
\centering
\includegraphics[scale=0.5]{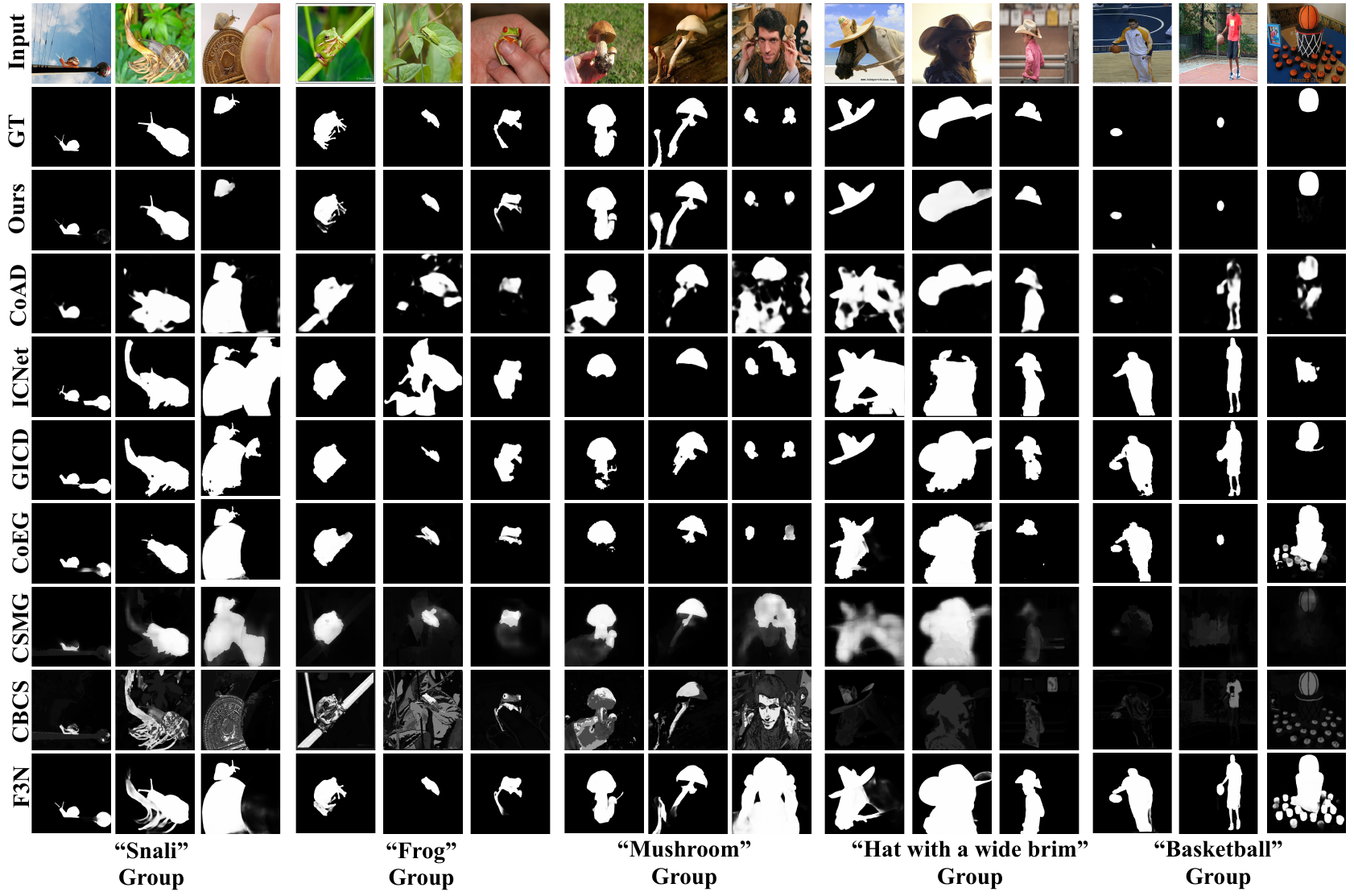}
\caption{Visual comparison between our method and other state-of-the-art methods.}
\label{Fig1}
\end{figure}

\begin{figure}[!t]
\centering
\includegraphics[scale=0.42]{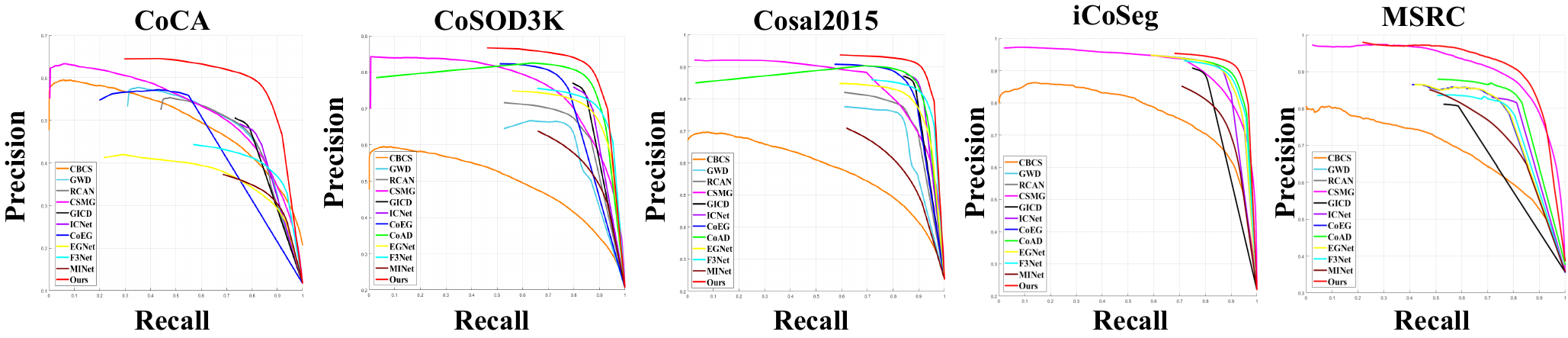}
\caption{Comparison of PR curves across five CoSOD datasets.}
\label{pr}
\end{figure}

As shown in Fig.\ref{Fig1}, we provide a comprehensive qualitative
comparison of our method with other state-of-the-art (SOTA) methods on challenging cases. When facing complex real-world scenarios, other methods are unable to handle these challenging cases, like "snail Group", "Hat Group" and "Basketball Group", where non-co-salient objects are very close to co-salient objects. While our proposed CoSformer models the pixel-level group relationships, and uses contrastive loss to model inter-image separability, therefore performs much better on detecting co-salient objects. As shown in Fig.\ref{pr}, we can see that our method (the red line) achieves the highest precision on all datasets. Our CoSformer runs averagely at 40 FPS on an Nvidia 2080Ti GPU. In conclusion, both qualitative and quantitative results in the main text and supplementary material demonstrate the superiority and effectiveness of our proposed CoSformer. 

\begin{table}[]
\caption{Quantitative comparison with SOTA methods on another two conventional datasets.}
\centering
\scalebox{0.85}{
\begin{tabular}{c|c|cccc|cccc}
\hline
                            &                        & \multicolumn{4}{c|}{iCoSeg}                                                                                               & \multicolumn{4}{c}{MSRC}                                                                                                  \\ \cline{3-10} 
\multirow{-2}{*}{Model}     & \multirow{-2}{*}{Type} & $E^{max}_{\phi}$             & $S_{\alpha}$                 & $F_\beta^{max}$              & MAE                          & $E^{max}_{\phi}$             & $S_{\alpha}$                 & $F_\beta^{max}$              & MAE                          \\ \hline
CBCS(TIP2013)               & Co                     & 0.797                        & 0.658                        & 0.705                        & 0.172                        & 0.676                        & 0.480                        & 0.630                        & 0.314                        \\
GWD(IJCAI2017)              & Co                     & 0.841                        & 0.801                        & 0.829                        & 0.132                        & 0.789                        & 0.719                        & 0.727                        & 0.210                        \\
RCAN(IJCAI2019)             & Co                     & 0.878                        & 0.820                        & 0.841                        & 0.122                        & 0.789                        & 0.719                        & 0.727                        & 0.210                        \\
CSMG(CVPR2019)              & Co                     & 0.889                        & 0.821                        & 0.850                        & 0.106                        & 0.859                        & 0.722                        & 0.847                        & 0.190                        \\
CoEG(TPAMI2020)             & Co                     & 0.912                        & 0.875                        & 0.876                        & 0.060                        & 0.793                        & 0.696                        & 0.751                        & 0.188                        \\
GICD(ECCV2020)              & Co                     & 0.891                        & 0.832                        & 0.845                        & 0.068                        & 0.726                        & 0.665                        & 0.692                        & 0.196                        \\
ICNet(NIPS2020)             & Co                     & 0.929                        & 0.869                        & 0.886                        & 0.047                        & 0.822                        & 0.731                        & 0.805                        & 0.160                        \\
CoAD(NIPS2020)              & Co                     & {\color[HTML]{32CB00} 0.930} & {\color[HTML]{32CB00} 0.878} & {\color[HTML]{32CB00} 0.889} & {\color[HTML]{32CB00} 0.045} & {\color[HTML]{32CB00} 0.850} & {\color[HTML]{32CB00} 0.782} & {\color[HTML]{32CB00} 0.842} & {\color[HTML]{32CB00} 0.132} \\
{\color[HTML]{000000} Ours} & Co                     & {\color[HTML]{FE0000} 0.943} & {\color[HTML]{FE0000} 0.904} & {\color[HTML]{FE0000} 0.907} & {\color[HTML]{FE0000} 0.038} & {\color[HTML]{FE0000} 0.869} & {\color[HTML]{FE0000} 0.795} & {\color[HTML]{FE0000} 0.852} & {\color[HTML]{FE0000} 0.122} \\ \hline
EGNet(ICCV2019)             & Sin                    & 0.911                        & 0.875                        & 0.875                        & 0.060                        & 0.794                        & 0.702                        & 0.752                        & 0.186                        \\
F3Net(AAAI2020)             & Sin                    & 0.918                        & 0.879                        & 0.874                        & 0.048                        & 0.811                        & 0.733                        & 0.763                        & 0.161                        \\
MINet(CVPR2020)             & Sin                    & 0.846                        & 0.789                        & 0.784                        & 0.099                        & 0.769                        & 0.688                        & 0.729                        & 0.194                        \\ \hline
\end{tabular}}
\label{table1}
\end{table}

\begin{table}[!hbt]
\caption{Ablation Studies of Losses}
\centering
\scalebox{0.9}{
\begin{tabular}{c|cccc|cccc}
\hline
                                 & \multicolumn{4}{c|}{CoSOD3k}                                                                                              & \multicolumn{4}{c}{Cosal2015}                                                                                             \\ \cline{2-9} 
\multirow{-2}{*}{Configurations} & $E^{max}_{\phi}$             & $S_{\alpha}$                 & $F_\beta^{max}$              & MAE                          & $E^{max}_{\phi}$             & $S_{\alpha}$                 & $F_\beta^{max}$              & MAE                          \\ \hline
CoAD(NIPS2020)                   & 0.874                        & 0.822                        & 0.786                        & 0.078                        & 0.915                        & 0.861                        & 0.857                        & 0.063                        \\
$Cont+BCE$                            & 0.878                       & 0.828                        & 0.800                        & 0.071                        & 0.921                        & 0.888                       & 0.884                        & 0.052                        \\
$Cont+BCE+SSIM$                       & {\color[HTML]{000000} 0.878} & {\color[HTML]{000000} 0.831} & {\color[HTML]{000000} 0.803} & {\color[HTML]{000000} 0.068} & {\color[HTML]{000000} 0.925} & {\color[HTML]{000000} 0.890} & {\color[HTML]{000000} 0.889} & {\color[HTML]{000000} 0.050} \\
$Cont+BCE+SSIM+F_m$                   & {\color[HTML]{FE0000} 0.879} & {\color[HTML]{FE0000} 0.835} & {\color[HTML]{FE0000} 0.807} & {\color[HTML]{FE0000} 0.066} & {\color[HTML]{FE0000} 0.929} & {\color[HTML]{FE0000} 0.894} & {\color[HTML]{FE0000} 0.891} & {\color[HTML]{FE0000} 0.047} \\ \hline
\end{tabular}}
\label{table2}
\end{table}

\begin{table}[]
\caption{Analyses of TGL module.}
\centering
\begin{tabular}{l|cccc|cccc}
\hline
\multicolumn{1}{c|}{}                                 & \multicolumn{4}{c|}{CoSOD3k}                                                                                              & \multicolumn{4}{c}{Cosal2015}                                                                                             \\ \cline{2-9} 
\multicolumn{1}{c|}{\multirow{-2}{*}{Configurations}} & $E^{max}_{\phi}$             & $S_{\alpha}$                 & $F_\beta^{max}$              & MAE                          & $E^{max}_{\phi}$             & $S_{\alpha}$                 & $F_\beta^{max}$              & MAE                          \\ \hline
Ours(order1)                                          & {\color[HTML]{000000} 0.879} & {\color[HTML]{000000} 0.835} & {\color[HTML]{000000} 0.807} & {\color[HTML]{000000} 0.066} & {\color[HTML]{000000} 0.929} & {\color[HTML]{000000} 0.894} & {\color[HTML]{000000} 0.891} & {\color[HTML]{000000} 0.047} \\
Ours(order2)                                          & {\color[HTML]{000000} 0.879} & {\color[HTML]{000000} 0.835} & {\color[HTML]{000000} 0.807} & {\color[HTML]{000000} 0.066} & {\color[HTML]{000000} 0.929} & {\color[HTML]{000000} 0.894} & {\color[HTML]{000000} 0.891} & {\color[HTML]{000000} 0.047} \\
Ours(order3)                                          & {\color[HTML]{000000} 0.879} & {\color[HTML]{000000} 0.835} & {\color[HTML]{000000} 0.807} & {\color[HTML]{000000} 0.066} & {\color[HTML]{000000} 0.929} & {\color[HTML]{000000} 0.894} & {\color[HTML]{000000} 0.891} & {\color[HTML]{000000} 0.047} \\ \hline
{\color[HTML]{6200C9} Ours(Std)}                      & {\color[HTML]{6200C9} 0.000} & {\color[HTML]{6200C9} 0.000} & {\color[HTML]{6200C9} 0.000} & {\color[HTML]{6200C9} 0.000} & {\color[HTML]{6200C9} 0.000} & {\color[HTML]{6200C9} 0.000} & {\color[HTML]{6200C9} 0.000} & {\color[HTML]{6200C9} 0.000} \\ \hline\hline
Ours-P(order1)                                        & {\color[HTML]{000000} 0.870} & {\color[HTML]{000000} 0.830} & {\color[HTML]{000000} 0.800} & {\color[HTML]{000000} 0.070} & {\color[HTML]{000000} 0.919} & {\color[HTML]{000000} 0.887} & {\color[HTML]{000000} 0.885} & {\color[HTML]{000000} 0.051} \\
Ours-P(order2)                                        & {\color[HTML]{000000} 0.878} & {\color[HTML]{000000} 0.835} & {\color[HTML]{000000} 0.807} & {\color[HTML]{000000} 0.066} & {\color[HTML]{000000} 0.928} & {\color[HTML]{000000} 0.893} & {\color[HTML]{000000} 0.891} & {\color[HTML]{000000} 0.047} \\
Ours-P(order3)                                        & {\color[HTML]{000000} 0.874} & {\color[HTML]{000000} 0.829} & {\color[HTML]{000000} 0.802} & {\color[HTML]{000000} 0.068} & {\color[HTML]{000000} 0.924} & {\color[HTML]{000000} 0.889} & {\color[HTML]{000000} 0.886} & {\color[HTML]{000000} 0.050} \\ \hline
{\color[HTML]{6200C9} Ours-P(Std)}                    & {\color[HTML]{6200C9} 0.005} & {\color[HTML]{6200C9} 0.003} & {\color[HTML]{6200C9} 0.004} & {\color[HTML]{6200C9} 0.002} & {\color[HTML]{6200C9} 0.004} & {\color[HTML]{6200C9} 0.004} & {\color[HTML]{6200C9} 0.003} & {\color[HTML]{6200C9} 0.002} \\ \hline\hline
RCAN(order1)                                          & 0.808                        & 0.744                        & 0.688                        & 0.130                        & 0.842                        & 0.779                        & 0.764                        & 0.126                        \\
RCAN(order2)                                          & 0.800                        & 0.732                        & 0.680                        & 0.138                        & 0.831                        & 0.764                        & 0.754                        & 0.136                        \\
RCAN(order3)                                          & 0.804                        & 0.739                        & 0.685                        & 0.132                        & 0.838                        & 0.772                        & 0.758                        & 0.130                        \\ \hline
{\color[HTML]{6200C9} RCAN(Std)}                      & {\color[HTML]{6200C9} 0.006} & {\color[HTML]{6200C9} 0.007} & {\color[HTML]{6200C9} 0.004} & {\color[HTML]{6200C9} 0.004} & {\color[HTML]{6200C9} 0.006} & {\color[HTML]{6200C9} 0.005} & {\color[HTML]{6200C9} 0.006} & {\color[HTML]{6200C9} 0.002} \\ \hline\hline
ICNet(order1)                                         & 0.832                        & 0.780                        & 0.743                        & 0.097                        & 0.900                        & 0.856                        & 0.855                        & 0.058                        \\
ICNet(order2)                                         & 0.827                        & 0.771                        & 0.736                        & 0.102                        & 0.893                        & 0.850                        & 0.845                        & 0.062                        \\
ICNet(order3)                                         & 0.825                        & 0.775                        & 0.739                        & 0.101                        & 0.896                        & 0.852                        & 0.850                        & 0.060                        \\ \hline
{\color[HTML]{6200C9} ICNet(Std)}                     & {\color[HTML]{6200C9} 0.007} & {\color[HTML]{6200C9} 0.005} & {\color[HTML]{6200C9} 0.003} & {\color[HTML]{6200C9} 0.004} & {\color[HTML]{6200C9} 0.005} & {\color[HTML]{6200C9} 0.004} & {\color[HTML]{6200C9} 0.008} & {\color[HTML]{6200C9} 0.003} \\ \hline
\end{tabular}
\label{TGL}
\end{table}

\section{Details of Losses}
As described in main text, Transformer can well model intra-image saliency and inter-image consistency. Contrast learning scheme can help model the inter-image separability and learn more discriminative representations to distinguish true common objects from noisy objects. Hence, as can be seen in Table.\ref{table2}, if we only supervise the predicted co-saliency maps $\mathcal{M}=\{M^{(n)}\}_{n=1}^N$ by the corresponding groundtruth $\mathcal{T}=\{{T}^{(n)}\}_{n=1}^N$ under  contrastive loss and BCE loss ($Cont+BCE$), the performance can already outperform the second ranked performance (COAD) by a large margin, which demonstrates the efficiency of the proposed CoSformer framework and contrast learning scheme. 

However, because of limited GPU memory , we only use the highest level feature $F^{(n)}_{6}$ ($8\times8$ resolution), which contains less detailed information, to model group-wise relationships. To make final predicted co-saliency maps contain more detailed information, we first use a simple U-shaped architecture that enables feature aggregation at low-level via skip-connections. Moreover, inspired by BASNet~\cite{DBLP:conf/cvpr/QinZHGDJ19}, which uses pixel-level, region-level and object-level supervision strategy to predict the salient objects with fine structures and clear boundaries, we also use $SSIM$
and $F_m$ losses in addition to BCE loss. The results can be seen in Table.\ref{table2}. A better performance has been achieved through the combination of $BCE$, $SSIM$ and $F_m$. The work~\cite{deng2020re} addresses that predicted co-saliency maps with fine boundaries is one of future directions, and we try to address this problem by simply using the $SSIM$ and $F_m$ losses in this paper. While the main contributions of this paper are the proposed CoSformer framework and contrast learning scheme. 

The equations of $BCE$,$SSIM$ and $F_m$ are shown below. It should be noted that, if the model is only trained with $Cont+BCE$ or $Cont+BCE+SSIM$, the $L_s$ (line.480 in main text) and $L_{ct}$ (line.516 in main text) only contain $BCE$ loss.

The BCE loss is defined as:
\begin{equation}
BCE = \sum_{n=1}^{N} -( T^{(n)}log(M^{(n)}) + (1-T^{(n)})log(1-M^{(n)})).
\end{equation}

Following the setting of~\cite{DBLP:journals/tip/WangBSS04,DBLP:conf/iccv/FanCLLB17}, we use the sliding window fashion to model region similarity between groundtruth and saliency map. The corresponding regions are denoted as $M_i^{(n)} = \{M_i^{(n)}:i=1,...D\}$ and $T_i^{(n)} = \{T_i^{(n)}:i=1,...D\}$, where $D$ is the total number of region. Then we use SSIM to evaluate the similarity between $M_i^{(n)}$ and $G_i^{(n)}$, which is defined as:
\begin{equation}
SSD_i^{(n)} = \frac{ (2\mu_m\mu_t+C_1)(2\sigma_{mt}+C_2)  }{ (\mu_m^2+\mu_t^2+C_1)(\sigma_m^2+\sigma_t^2+C_2)}
\end{equation}
where local statistics $\mu_m$, $\sigma_m$ is mean and std vector of $S_i^{(n)}$, $\mu_t$, $\sigma_t$ is mean and std vector of $T_i^{(n)}$. The overall loss function is defined as:
\begin{equation}
SSIM = \sum_{n=1}^{N} (1 - \frac{1}{D}\sum_{i=1}^D SSD_i^{(n)}).
\end{equation}

Finally, inspired by~\cite{DBLP:conf/iccv/ZhaoGWC19}, we directly optimize the F-measure to learn the global information from groundtruth. For easy remembering, we denote F-measure as $F_\beta$ in the following. $F_\beta^{(n)}$ is defined as:
\begin{equation}
precision^{(n)} = \frac{ \sum{ M^{(n)} \cdot T^{(n)} }}{ \sum{M^{(n)}} + \epsilon}, \ \ recall^{(n)} = \frac{ \sum{ M^{(n)} \cdot T^{(n)} }}{ \sum{T^{(n)}} + \epsilon},
\end{equation}
\begin{equation}
F_\beta^{(n)} = \frac{ (1+\beta^2) \cdot precision^{(n)} \cdot recall^{(n)} }{ \beta^2 \cdot precision^{(n)} + recall^{(n)} },
\end{equation}
where $\cdot$ means pixel-wise multiplication, $\epsilon=1e^{-7}$ is a regularization constant to avoid division of zero. $L_{Object}$ loss function is defined as:
\begin{equation}
F_m = \sum_{n=1}^{N} (1 -  F_\beta^{(n)}).
\end{equation}

\section{Analyses of TGL module}
In main text, we claim that our proposed TGL can let the CoSformer insensitive to the input order of group images, which can greatly improve the stability of CoSOD network. We further verify this through experiments, and the results are shown in Table.\ref{TGL}. Specifically, during testing, for each image group, we randomize 10 different orders, and only show three results (Ours) in Table.\ref{TGL} because of limited space. It can be seen that performance has no change.
{\color[HTML]{6200C9} Ours(Std)} means the standard deviation of these 10 orders. This result verifies that our proposed TGL can let the CoSformer insensitive to the input order of group images. Moreover, we do additional experiments to see what the impact would be when the positional encoding is added to the TGL module during training and testing. During testing, for each image group, we randomize 10 different orders, and show three results (Ours-P) in Table.\ref{TGL} as representations. It can be seen that extra positional encoding information will destabilize the CoSformer framework. Positional encoding assigning input order to related images, if the order of an image changed, the output group representation from TGL will be different. So adding positional encoding in TGL would make inter-image relationship modeling sub-optimal.

In the Introduction of the main text, we argue that existing sequential order modeling approaches make CoSOD networks unstable. So we do experiments on two typical methods, including RCAN~\cite{DBLP:conf/ijcai/0061STSS19} and ICNet~\cite{jin2020icnet}, to verify the inferring procedure of these two methods are unstable. Because CoAD~\cite{DBLP:conf/nips/ZhangCHLZ20} does not release their code, so we can not do experiments on CoAD. During testing, for each image group, we randomize 10 different orders, and the three results are shown in Table.\ref{TGL}. As can be seen in Table.\ref{TGL}, whether using RNN (RCAN), or applying some sophisticated modification on CNNs architectures (ICNet), can not let the CoSOD network insensitive to the input order of group images, leading to an unstable inferring procedure. Because in sequential order modeling, both CNNs and RNNs have inherent deficiencies. Through these experiments, we further verify that the proposed CoSformer  greatly improves the stability CoSOD network.

{\small
\bibliographystyle{ieee_fullname}
\bibliography{references}
}

\end{document}